\documentclass[11pt,a4paper]{article}
\usepackage[hyperref]{acl2021}
\usepackage[whole]{bxcjkjatype}
\usepackage{graphicx}
\usepackage{color}
\usepackage{booktabs}
\usepackage{multirow}
\usepackage{colortbl}
\usepackage{times}
\usepackage{latexsym}
\usepackage{subcaption}
\usepackage{amsmath,amssymb}

\usepackage{microtype}
\newcommand{\tabref}[1]{Table \ref{#1}}

\newcommand{\figcaption}[1]{\def\@captype{figure}\caption{#1}}
\newcommand{\tblcaption}[1]{\def\@captype{table}\caption{#1}}
\aclfinalcopy 

\title{Empirical Analysis of Training Strategies of Transformer-based Japanese Chit-chat Systems}

\author{Hiroaki Sugiyama \\\And
  Masahiro Mizukami \\\And
  Tsunehiro Arimoto \\\And
  Hiromi Narimatsu \\\AND
  Yuya Chiba \\\And
  Hideharu Nakajima \\~\\NTT Communication Science Laboratories, Japan \\\And
  Toyomi Meguro \\
  }

\date{}

\begin{document}
\maketitle
\begin{abstract}
In recent years, several high-performance conversational systems have been proposed based on the Transformer encoder-decoder model.
Although previous studies analyzed the effects of the model parameters and the decoding method on subjective dialogue evaluations with overall metrics, they did not analyze how the differences of fine-tuning datasets affect on user's detailed impression. 
In addition, the Transformer-based approach has only been verified for English, not for such languages with large inter-language distances as Japanese.
In this study, we develop large-scale Transformer-based Japanese dialogue models and Japanese chit-chat datasets to examine the effectiveness of the Transformer-based approach for building chit-chat dialogue systems.
We evaluated and analyzed the impressions of human dialogues in different fine-tuning datasets, model parameters, and the use of additional information.

\end{abstract}

\section{Introduction}
A large-scale Transformer encoder-decoder model, which reportedly has high performance in question answering and machine translation in the adjacent fields of dialogue \cite{wang2018glue,wang2019acl}, has been applied to a chat dialogue system \cite{Adiwardana2020TowardsChatbot,stephen2020blenderbot}.
There are two chit-chat dialogue systems based on the Transformer model: Meena by Google and BlenderBot by Facebook.
Both use for pre-training a large number of response pairs that are collected from SNS responses or discussions on bulletin boards.
Meena is the first neural-based system that clearly surpasses the conventional rule-based systems. It only uses simple response pairs obtained from SNS and trains them with an extremely large amount of computation.
BlenderBot is a large-scale, transformer, encoder-decoder model pre-trained with Reddit data and fine-tuned with a corpus of three skills (personality, empathy and knowledge) defined by the authors as necessary for chatting dialogues. 
By training on high-quality data, BlenderBot subjectively outperformed Meena with the same model size with less computation and shows similar performance in comparison with humans.

\begin{table}[t]
    \centering
    \footnotesize
    \begin{tabular}{r|p{20em}}
System& Let's talk about your favorite things. (早速ですがご趣味をうかがってもいいですか?)\\
User& I like traveling, especially traveling alone. (私の趣味は、旅行です！一人旅！)\\
System& Awesome! Do you generally travel domestically or overseas? (そうなんですね! 旅行は国内ですか?海外ですか?)\\
User& Overseas. I love going to Europe! (海外旅行のほうが多いです！ヨーロッパとか大好きなんです)\\
System& Europe sounds great. Which country have you been to? (ヨーロッパいいですね! 特にどの国に行かれましたか?)\\
User& Italy, many times to Venice. (何度も行ってるのは、イタリアのベネチアです)\\
System& I've never been to Venice. What's it like? (ベネチア行ったことないです! どんなところですか?)\\
User& It's a floating city, and the whole city is World Heritage. We can travel around by boats. (街全体が世界遺産の水上都市です。交通手段が船だけなんですよ)\\
System& That's interesting. I want to visit there. The scenery looks really beautiful! (えー!すごいですね! 行ってみたいです! 景色も綺麗そうですね!)\\
User& Exactly! I took a lot of pictures. (まさに！景色も最高です！写真撮りまくりです！)\\
System& I like taking pictures, too. What kind do you take? (写真撮るの好きです! どんなジャンル撮られるんですか?)\\
User& I often take pictures of scenery when traveling. (写真楽しいですよね！私は旅行先で景色撮ることが多いです)\\
System& So do I. Have you traveled anywhere recently? (私も同じです。旅行先で景色撮ってます。 最近どこか行かれましたか?)\\
User& No, not really, although I did recently go to Kyoto. (最近はなかなか出かけてないですが、少し前に京都行きましたよ)\\
    \end{tabular}
    \caption{Example of Japanese chit-chat with our Transformer-based model (trained with our favorite-things dialogue dataset)}
    \label{tab:hobbyist}
\end{table}

BlenderBot showed good performance in generating chit-chat responses. Although its model behaviors have been analyzed, three remaining unrevealed issues have not yet been addressed.
The first issue is how the characteristics of the training corpus affect user impressions.
Since conventional work examined only one setting of the fine-tuning datasets (even mixed datasets) of their models and just used a few overall evaluation metrics (e.g., ACUTE, SSA), the effects of varying the fine-tuning datasets on user impressions have not been examined.

Another problem is that the system's effectiveness has only been verified for English, not for such languages with large inter-language distances as Japanese.
Due to the differences in language resources and community sizes, non-English initiatives on pre-trained models are much less common compared to English.
Multilingual BART (mBART) is one initiative that has used a multilingual corpus for simultaneous learning in multiple languages \cite{liu-etal-2020-multilingual-denoising}.
Although it works well for languages with close linguistic characteristics, such as European languages, it has performed less favorably for languages with distant linguistic characteristics, such as Japanese (especially in the generation task).

In this study, we develop large-scale Transformer-based Japanese dialogue models and Japanese chit-chat datasets to examine the effectiveness of the Transformer-based approach for building chit-chat dialogue systems.
We also analyze the relationship between user impressions and the fine-tuning strategies of the Transformer model, such as the dataset characteristic and amount, model size, and the presence or absence of additional information.
Since we expect the above fine-tuning strategies to affect various impressions, this study established multiple evaluation scales to conduct a multifaceted evaluation of the Transformer-based chatting system.

The following are the contributions of this paper:
\begin{itemize}
\item Pre-trained, evaluated, and published a Japanese dialogue model with data comparable in scale to the SoTA systems in English\footnote{We are preparing for releasing the models and datasets.} (Table \ref{tab:hobbyist}).
\item Created and published benchmark data for a chit-chat dialogue system in Japanese.
\item Analyzed the impact of fine-tuning strategies (datasets used, model size and the use of additional information) on subjective evaluations.
\end{itemize}

\begin{table*}[t]
    \centering
\scalebox{1.0}[1.0]{
    \begin{tabular}{rrrrrrrrr}
        Model name & Total Parameters & $V$ & $L_{enc}$ & $L_{dec}$ & $d$ & $h$ &Steps & PPL\\\hline
        0.35B & 359,607,808 & 32K & 2 & 22 & 896 & 32 & 46K & 5.159 \\
        0.7B & 698,536,960 & 32K & 2 & 22 & 1280 & 32 & 48K & 5.033 \\
        1.1B & 1,065,683,200 & 32K & 2 & 22 & 1600 & 32 & 48K & 4.968 \\
        1.6B & 1,627,872,000 & 32K & 2 & 24 & 1920 & 32 & 48K & 4.924 \\\hline
   \end{tabular}
    }
    \caption{Training parameters of pre-trained models and perplexity on the validation set of our Twitter pre-training dataset for several models with given architecture settings. Columns include the vocabulary size (V), number of encoder and decoder layers ($L_{enc}$, $L_{dec}$), embedding dimensionality (d), Multihead Attention Heads (h), and training steps.}
    \label{tab:pre-train}
\end{table*}

\section{Pre-training}
\subsection{Model architecture}

We generated responses with standard Seq2Seq Transformer encoder-decoder architecture, which was also used with the {\it generative} model of BlenderBot \cite{stephen2020blenderbot}.
We used Sentencepiece tokenizer \cite{kudo-richardson-2018-sentencepiece} as implemented in the Official Github site \footnote{https://github.com/google/sentencepiece} to avoid unnecessary dependency on a specific fixed vocabulary.

We examined the improvement in model size in detail by considering four model sizes: 0.35B, 0.7B, 1.1B, and 1.6B parameters.
Although our largest model (1.6B parameter) is slightly smaller than the original BlenderBot 2.7B because of the limitation of our computation resources, we believe that the 1.6B-sized model is sufficient to examine the effect of model size.

\subsection{Pre-training dataset}

For standard sentence classification or machine translation, plain text is commonly used for pre-training with denoising tasks that aim to recover original sentences from noise-added ones.
Meena and BlenderBot, on the other hand, use a large amount of interactive data extracted from social networking sites or Reddit as pre-training to learn the relationship between direct input contexts and target utterances.

We follow the conventional research and utilize a large amount of Twitter reply pairs, which are interactive data, for pre-training.
The following is our data cleaning and setup procedures.
First, we retrieved all the tweets from January 2016 to March 2018 of randomly sampled Japanese users. 
After a tweet-cleaning process, we performed declension and removed the account names and emojis. Then we removed the tweets that match the following conditions:

\begin{itemize}
  \setlength{\parskip}{0cm} 
  \setlength{\itemsep}{0cm} 
\item Tweets that have another tweet with a cosine similarity of 0.9 or higher on the same day (tweets with fewer than 20 characters are not filtered).
\item Retweets.
\item Tweets that contain URLs.
\item Tweets that contain parentheses to prevent emoticons.
\item Tweets where the user is a bot.
\item Tweets that contain fewer than 30\% Hiragana and Katakana characters.
\end{itemize}

Next we extracted the tweets in a reply relationship from the cleaned tweets and paired them with the input contexts and target utterances.
Using the tweet pairs, we extracted reply chains of tweets by extending the chain one by one from its root tweet.
We utilized the last tweet of a chain as a target utterance and the rest that contain the root as the input context.
For example, if the chain is A-B-C-D, we used A-B, AB-C, and ABC-D, but not B-C.
 We obtained 2.1 billion (521 GB) pairs by this method.
The average number of utterances in the input context was 2.91\footnote{The input context was calculated using 0.12\% of the total data.}, and the average number of characters was 62.3 for the input context and 20.3 for the target utterance.
We built a Sentencepiece tokenization model \cite{kudo-richardson-2018-sentencepiece} using 20 million sentences sampled from the data of a QA community service called "Oshiete goo!\footnote{https://oshiete.goo.ne.jp/}" since the data cover the period from 2001 to 2019 and contain more recent topics than our pre-training data.

\subsection{Training details}

The model parameters are based on the 2.7 billion parameters of Meena and BlenderBot, where the encoder is two layers and the decoder is 22 or 24 layers.
The number of dimensions of the hidden layers is adjusted to avoid memory errors on the GPU (V100 16GB) available at AIST ABCI Cloud\footnote{https://abci.ai/ja/}, which is the computing resource we used.
\tabref{tab:pre-train} shows the training parameters of each pre-training model used in this study, which are related to the model size.

The other parameters are explored using Weight and Biases \cite{wandb} and set as follows.
The dropout of the feed-forward layer and attention is set to 0.1.
The learning rate is 1e-3, with 10000 warmup steps, and the maximum number of tokens per step is 2.4M.
The objective function is set to label the smoothed cross entropy to promote early learning.
Our computational resources were 400 V100 16GB cards.
48000 was the maximum number of training steps for the 1.6B model, where the early stopping steps of the training were about 45000 steps, which is almost equivalent to three epochs.
The input format to the encoder connected the utterances in the input context of each pair with [SEP] tokens; no other information was added.

The implementation uses the TransformerEncoderDecoderModel from fairseq\footnote{https://github.com/pytorch/fairseq} that was trained on a translation task.
When we tried to pre-train on a single dataset, the data were too large. Therefore, we grouped the reply pairs by their Tweet dates and trained them by data sharding.
The validation data were set to the 3/28/2018 data.

\section{Fine-tuning}
\label{sec:fine-tuning}
In this study, we created a Japanese version of a similar corpus using BlenderBot as a reference and used it for fine-tuning.
In this section, we describe the corpus used for fine-tuning, the format of the input information, and the detail settings of the fine-tuning.

\subsection{Fine-tuning dataset}


For fine-tuning BlenderBot, \citet{stephen2020blenderbot} used PersonaChat, EmpatheticDialogues, and Wizard of Wikipedia as datasets that individually correspond to three abilities: personality, empathy, and knowledge, which should be possessed by a chit-chat system, and simultaneously used BlendedSkillTalk dataset to integrate the abilities.
For fine-tuning our Japanese version of BlenderBot, we develop a Japanese version of PersonaChat and EmpatheticDialogues and our original FavoriteThingsChat datasets.
Although we also tried to construct a Wizard of Wikipedia, conducting meaningful conversations was actually very difficult.
In the construction of Wizard of Wikipedia dataset, a knowledge-offering interlocutor (wizard) of each dialogue can refer only the first paragraph of Wikipedia pages, which gives a definition of the page content and is insufficient for the wizard to expand the dialogue meaningfully.
Although we examined the translation from the original English Wizard of Wikipedia to Japanese one, many topics were different from those that generally appear in Japanese conversations. After determining that it did not contribute to the learning, we abandoned the utilization of Wizard of Wikipedia.


The unavailability of the Wizard of Wikipedia greatly complicated building the BlendedSkillTalk, which requires dialogue models that are learned in each of the three datasets.
As an alternative of BlendedSkillTalk, we originally develop FavoriteThingsChat dataset, which also contains utterances displaying empathy, knowledge and consistent personality.
The details of each corpus are described below.



\subsubsection{PersonaChat（PC）}

PersonaChat \cite{zhang2018acl} is a corpus where each speaker sets five profile sentences that regulates the features of the speaker.
Conversations are conducted based on the given profile sentence set, and the conversations of various speakers are collected in a pseudo manner.
In this study, we constructed a Japanese version corpus of Persona-chat by creating a Japanese version of a profile sentence set and collecting conversations by Japanese speakers.

The Japanese version of the profile sentence set is made as one profile sentence set by combining five sentences of 30 characters or fewer following a previous method \cite{zhang2018acl}.
Cloud workers created 100 sets.
A trained annotator rewrote some sentences to remove similar structures or words.

In the dialogue collection, 100 sets of obtained profile sentences were allocated to each cloud worker, and 5000 dialogues were collected.
All the cloud workers engaged in chats to talk about the content of the set of profile sentences given to them.
Each utterance alternately carried out one utterance, and a dialogue was collected so that one utterance can consist of a maximum of 100 characters, a minimum of 12 utterances, and a maximum of 15 utterances (6-8 turns).
61794 utterances were included in the 5000 collected dialogues.

\subsubsection{EmpatheticDialogues（ED）}
EmpatheticDialogues is a dataset that collects dialogues with an open-domain one-on-one conversational setting where two people are discussing a situation that happened to one of them, related to a given emotion \cite{rashkin2019empathetic}.
\citet{rashkin2019empathetic} used crowdsourcing to collect 24,850 dialogues that are grounded descriptions of situations in which a speaker expresses a given feeling of 32 emotional words.
In this study, to make a Japanese version of EmpatheticDialogues, we translated 32 English words that show emotions into Japanese, and a Japanese speaker used them to construct situation sentences and dialogues.
To reduce collection cost, one dialogue was not an actual dialogue done by two workers; it was a pseudo dialogue written by one worker.
The crowdworker refers to the translated list of emotions and creates a context sentence of 1-3 sentences based on the emotions and a text dialogue of four utterances by two persons (speaker and listener) who interact in the context.
20,000 dialogues and 80,000 pairs of utterances were collected.

\subsubsection{FavoriteThingsChat dataset（Fav）}

The FavoriteThingsChat dataset consists of intensively collected chats about the favorites of various people.
All 80 experiment's participants talked with more than 60 other participants.
We collected 3480 dialogues and 123,069 utterances.
Since the participants talk about their own actual favorites, they proactively show knowledge about their own favorites and display empathy and interest about the interlocutor's favorites with consistent personality. 
The range of the presented topics is comparatively narrow, because they are limited to the favorite things of each speaker, and because only 80 speakers repeatedly talk with each other participants. 
We expect that dialogues collected with such the "high density" dialogue collection setting are helpful for the dialogue model to have enough knowledge to speak each dialogue topic deeply.
In addition, each conversation continues for a relatively large number of turns (average of 35.3), which is a suitable setting for learning a long conversation.
We expect that the learning will improve the dialogue impressions more than those conducted by more PersonaChat where the number of turns is low and the speaker plays the Persona and Empathetic Dialogues with fewer turns and much data are deleted from one dialogue scene.
\tabref{tab:favdial} shows an example of the collected dialogues.

\begin{table}[t]
    \centering
    \footnotesize
    \begin{tabular}{r|p{20em}}
Speaker & Utterance \\\hline
65 & Hello! (こんにちは)\\
71 &(Pleased to meet you! (よろしくお願いします！)\\
67 &What do you do in your free time? (趣味はなんですか？)\\
71 &I like traveling, watching movies and reading books! How about you? (私の趣味は、旅行と映画を観たり本を読んだりすることです！ あなたの趣味は何ですか？)\\
67 & I like watching baseball games and playing video games Where have you traveled to recently? (趣味は野球観戦とゲームですねー　旅行は最近どちらに行かれましたか？)\\
71& Do you have a favorite team to watch baseball games? (野球観戦は好きなチームがありますか？)\\
67& For professional baseball, I like Orix. (プロ野球だとオリックスですねー)\\
71& Recently, I went to Okinawa last July and Toyama in May! Orix is the team where Ichiro joined, right? (最近は去年の7月に沖縄と5月に富山に行きました！オリックスは昔イチローがいたチームですね？)\\
67&Yes! But when I started rooting for them, he was in the major leagues... What kind of places did you visit in Toyama...?(そうです！ただ僕が応援し始めたときにはメジャーリーグに行っていましたねー 富山はどういったところ廻られたんですか…？) \\
71& That was a long time ago! Do you go to see games at the stadium? (結構昔ですもんね！！！試合をドームとかに観に行ったりもされるんですか？)\\
67 & Yes, I do. I didn't go to many games last year, but I think I went to about 10 games the year before last. (行きますねー 去年はあんまりでしたが、一昨年は10試合ほど行ったと思いますー)\\
71 & In Toyama, there is a park with tulips, the most beautiful Starbucks in Japan, and a glass museum. I went to a park with tulips, a beautiful Starbucks and a glass museum! I also ate white shrimps and Toyama black (ramen)! (富山は、チューリップがある公園と、日本一？美しいスタバとガラスの美術館に行きました！ あとは白エビ食べたり、富山ブラック（ラーメン）食べたりしました！)\\
    \end{tabular}
    \caption{Example of our FavoriteThingsChat dataset)}
    \label{tab:favdial}
\end{table}

\subsubsection{Mixed dataset（Mix）}
In BlenderBot's generative model, several kinds of corpus are mixed for fine-tuning.
No research has clarified yet whether the interaction function, which is the intention of each dataset, is achieved.
Nor has research investigated which improves the interaction performance more: a high-quality dataset that is used singly or when the overall quantity is increased by adding another dataset.
In this study, in addition to the case in which the above dataset is used for the fine-tuning alone, we compare the following two cases: one mixed in the same quantity as each dataset and another where each dataset was mixed in the whole quantity.

\subsection{Use of additional information in query sentences}

\label{sec:addinfo}

In response generation using the encoder-decoder model, when information is input to the encoder,
in addition to the dialogue’s context, additional information can also be input in the same text format.
In this study, we analyzed the effect of the presence or absence of such additional information on the impressions of dialogues.
Such information might improve the generation performance because 
clinging to it deleteriously affects long dialogues.

Below we show the information to be added to each dataset.
PersonaChat, as in a previous work, adds a set of profile statements to the input \cite{zhang2018acl} to
improve the utterance consistency by generating responses by linking them.
In EmpatheticDialogues, situation sentences and emotional words are added to input sentences, as in previous studies. 
The stability of utterance generation is expected to increase since the situation and the feeling to be generated are decided \cite{rashkin2019empathetic}.
In the FavoriteThingsChat, only the speaker's ID is added as information. In comparison with the above two datasets of PersonaChat and EmpatheticDialogues, the effect seems comparatively small, because the information is not accompanied by concrete content in a simple ID.

\begin{table}[h]
    \centering
    \small
    \begin{tabular}{p{\linewidth}}
    $<${\it Dataset name}$>$:[SEP]$<${\it Speaker ID}$>$[SEP][SPK1] $<${\it System Utt.}$>_{t-2}$[SEP][SPK2] $<$ {\it User Utt.}$>_{t-1}$ 
    \end{tabular}
    \caption{Query sentence format input for Encoder}
    \label{tab:input_template}
  \end{table}

\subsection{Fine-tuning training details}
In fine-tuning, up to four utterances are used as a dialogue context until the maximum character length reaches 128.
As in the pre-training, Adafactor was used as the optimizer for training.
The other parameters were changed from the pre-training settings: the learning rate was 1e-4, 3000 warmup steps, and a batch size of 256.
With these settings, we trained up to 3000 steps (about 30 minutes with 128 V100 16GB cards) with a model that minimized the perplexity of the validation set.

\section{Sentence generation settings}
\paragraph{Decoding}

\label{sec:decode}

For decoding the utterances from the model, we adopted the sample-and-rank format as in Meena \cite{Adiwardana2020TowardsChatbot}.
In this method, the final output is the candidate with the lowest perplexity among $N$ speech candidates generated independently by sampling.
In our initial study, we used the diverse beam search method that resembles BlenderBot. However, we found that the sample-and-rank method was more advantageous for expanding the topics because the diverse beam search often produced somewhat boring responses.

We also introduced temperature $T$ when calculating the softmax that is used for controlling the token output probability \cite{hinton2015distilling}. 
A temperature of $T=1$ results in normal sampling, and the higher temperature $T$ is, the more contextually unusual tokens (e.g., proper nouns) are generated. At the same time, contextually incorrect tokens are more likely to be generated.
Conversely, the lower temperature $T$ is, the more likely safe and common words will be selected.
In addition, we used nucleus sampling to limit the number of words sampled by the probability cumulative density \cite{holtzman2019topp}.
We used $top\_p=0.9$ and $T=1.0$ based on preliminary experiments.

\paragraph{Filtering candidate utterances}

In a preliminary experiment, we found that many repetitive utterances, generated from the models, have almost identical content as the utterances in the past context.
To suppress such repetitive utterances, we filtered the candidate utterances with the similarity of the Gestalt pattern matching algorithm\footnote{https://docs.python.org/ja/3/library/difflib.html} with utterances in the context and sentences (segmented by punctuation from the context utterances) exceed threshold $\sigma_r$.
We set $\sigma_r$ to 0.5.

\begin{table*}[t]
\small
    \centering
    \begin{tabular}{r|l}
Metric name & Questionnaire \\\hline
Humanness & The system utterances were human-like and natural. \\ &(システムの発話は人間らしく自然だった)\\
Ease & Continuing the dialogue was easy. \\&(簡単に対話を続けることができた)\\
Enjoyability & I enjoyed interacting with the system.\\ &(システムとの対話は楽しかった)\\
Empathetic&I was able to empathize with the system utterances.\\& (システムの発話に共感できた)\\
Attentiveness & The system was interested in me and was actively trying to talk with me.\\&(システムはあなたに興味を持って積極的に話そうとしていた)\\
Trust & I felt that what the system said was trustworthy.\\ &(システムの話したことは信頼できると感じた)\\
Personality & I could sense the system's personality and character.\\&(システムの個性・人となりが感じられた)\\
Agency & I felt that the system was speaking from its own perspective.\\&(システムは自身の考えをもって話していると感じた)\\
Topic & I felt that the system had a topic it wanted to discuss.\\&(システムには話したい話題があると感じた)\\
Emotion & I felt that the system had feelings.\\&(システムは感情を持っていると感じた)\\
Consistency&The system utterances were consistent and coherent.\\&(システムの発話は矛盾せず一貫していた)\\
Involvement&I was absorbed in this dialogue.\\&(この対話にのめりこめた)\\
Respeak&I want to talk with this system again.\\&(またこのシステムと話したい)\\\hline
\end{tabular}\vspace{-0mm}
    \caption{Evaluation metrics}\vspace{0mm}
    \label{tab:criteria}
\end{table*}

\section{Experiment}
As described in Section \ref{sec:fine-tuning}, we analyzed how the user's impressions of the dialogue changed depending on the dataset used for fine-tuning and inputting additional information to the encoder.
We also analyzed the effects of the mixture of datasets and the model size on the overall performance.

\begin{table*}[t]
    \centering
    \begin{tabular}{ccccccc}
Fine-tuning corpus & Size & PC & ED & Fav & Mix\\\hline\hline
Pre-trained & 1.6B & 38.45/39.50 & 27.35/28.03 & 29.35/33.41 & 31.65/33.67 \\\hline
PC50k & 0.35B & 25.03/21.72 & 27.83/21.89 & 39.57/35.2 & 30.29/25.75 \\
PC50k & 0.7B & 23.06/19.77 & 24.11/19.41 & 35.25/31.30 & 27.08/23.07 \\
PC50k & 1.1B & 21.88/18.86 & 22.89/18.06 & 34.71/30.42 & 26.03/21.99 \\
PC50k & 1.6B & 21.32/18.35 & 22.15/17.58 & 34.06/29.58 & 25.38/21.39 \\\hline
ED50k & 0.35B & 42.84/33.92 & 19.72/15.64 & 38.64/37.05 & 32.86/27.82 \\
ED50k & 0.7B & 39.15/30.50 & 17.81/14.13 & 35.99/34.09 & 30.12/25.25 \\
ED50k & 1.1B & 38.47/28.78 & 16.97/13.42 & 35.53/33.39 & 29.37/24.19 \\
ED50k & 1.6B & 34.22/28.26 & 16.21/13.05 & 31.05/32.26 & 26.52/23.54 \\\hline
Fav50k & 0.35B & 44.97/42.13 & 31.37/27.48 & 21.74/21.07 & 31.41/29.19 \\
Fav50k & 0.7B & 41.50/39.34 & 28.46/25.12 & 19.97/19.60 & 28.79/27.05 \\
Fav50k & 1.1B & 39.83/35.91 & 26.85/23.11 & 19.12/18.54 & 27.47/25.05 \\
Fav50k & 1.6B & 37.23/34.79 & 25.26/22.21 & 19.05/17.94 & 26.30/24.21 \\\hline
Mix50k & 0.35B & 28.91/24.3 & 21.43/17.15 & 23.25/23.11 & 24.53/21.55 \\
Mix50k & 0.7B & 26.27/22.00 & 19.23/15.43 & 21.36/21.20 & 22.29/19.56 \\
Mix50k & 1.1B & 25.04/21.01 & 18.24/14.57 & 20.35/20.23 & 21.22/18.61 \\
Mix50k & 1.6B & 24.21/20.43 & 17.58/14.20 & 19.83/19.60 & 20.55/18.09 \\\hline
Mix150k & 0.35B & 25.64/21.84 & 20.10/15.91 & 22.19/21.54 & 22.69/19.8 \\
Mix150k & 0.7B & 23.52/20.00 & 18.00/14.33 & 20.48/20.02 & 20.71/18.13 \\
Mix150k & 1.1B & 22.35/19.04 & 17.04/13.50 & 19.53/19.00 & 19.68/17.19 \\
Mix150k & 1.6B & 21.69/18.46 & 16.41/13.09 & 18.94/18.24 & 19.05/16.61 \\\hline
        \end{tabular}
    \caption{Perplexity of compared models on each test dataset. Left values show flat (no additional information) condition and right show tagged (with additional information) condition.}
    \label{tab:ppl}
\end{table*}

\subsection{Experiment procedures for verification}
This section explains our verifying factors for fine-tuning and experiment procedures.

\paragraph{Fine-tuning datasets}
The utterances contained in each dataset have different properties depending on the dialogue function intended by the dataset.
For example, EmpatheticDialogues are expected to have empathetic and emotional utterances, and PersonaChat to have questions and self-disclosing utterances about the interlocutors' personalities.
These properties will give users different dialogue impression.
We analyze how the user's impression of the dialogue with the model changed depending on the nature of the utterances in the fine-tuning datasets.

First, we train a dialogue model using only the utterances contained in each dataset, without additional information such as profile sentences.
We call this setting as {\it flat}.
Then we let the experiment participants interact with all the model and evaluate each dialogue using 13 measures described below.
We compare the average values of the 13 metrics among the models $v(m^{flat}_d, s) \in D$ to verify overall performance of the models.
We also compare the values of the 13 metrics with their averages for each fine-tuning dataset to verify whether each contributes to the value of a particular scale. 
Note that, since we expected the range of values for each metric to be different, we calculate normalized values of each metric with subtracting the average value of each metric for each dataset from the values assigned to each metric 
to reduce the effect of any differences in the metrics themselves.
If the difference value of a metric has a large absolute amount, corpus-specific effects on the metric are being observed.
We performed Friedman test for repeated measurements \cite{friedman1937test} on the differences of the normalized value of each metric for each dataset.
For the dataset that are found to be significant, we perform the Wilcoxon signed rank test \cite{wilcoxon1945signed} to examine the difference between each metric and the averaged scores.
For the multiple testing correction, we adopt the BH procedure that controls False Discovery Rate \cite{benjamini1995fdrbh}.

\paragraph{Using additional information}
We analyzed the effect of the presence of additional information on the dialogue impressions based on Section \ref{sec:addinfo}.
Even though using additional information may improve the generation performance, it might also negatively impact long dialogues because of adherence to the additional information.
We verify the effect of the additional information through almost the same process as the {\it flat} condition described above, with the difference of using additional information in the fine-tuning ({\it tagged} condition).

\paragraph{Mixing fine-tuning datasets}

We investigated the impact of mixing multiple datasets on the overall model performance.
We considered two methods.
The first trained the model on the same amount of data as the individual datasets to test the usefulness of mixing different types of dialogues.
Although training on a wide range of data might improve the overall performance by robustly training the model, when mixing high- and low-quality datasets, the performance might only be improved with a high-quality dataset.
The second case simply increases the amount of data. In this case, we examined whether the performance is improved by increasing the amount of data, even with low-quality data.

In addition, we mixed the datasets under two conditions: one adds a word that represents each corpus ("個性雑談" (meaning PersonaChat in Japanese), "共感雑談" (EmpatheticDialogues), and "趣味雑談" (FavoriteThingsChat))) at the beginning of each input sentence of the dataset and additional information that corresponds to each dataset (we call this as {\it mixed-tagged} condition); the other only fine-tunes from utterances without any additional information ({\it mixed-flat}). 
In the inference for actual conversations of mixed-tagged condition, we use dataset type "趣味雑談" (FavoriteThingsChat)) and randomly set IDs to be added based on the settings of FavoriteThingsChat to minimize the additional information.

\paragraph{Model size and fine-tuning datasets}
In BlenderBot, the performance did not improve even when the model size increased.
We investigated the effect of varying the model size and the training dataset on the performance.
We used the average value of each measure and examined whether the evaluation values correlated with the model size for each dataset.

\subsection{Evaluation metrics}
\label{sec:criteria}
\citet{fitrianie2020iva} conducted an extensive meta-survey of evaluation measures of interactive virtual agents and user interaction and classified those used in existing research. 
The classified evaluation measures contain various perspectives, and are useful references for designing evaluation metrics for dialogue systems in our study. However, since the scales are not for textual dialogues but for dialogues with CG-visualized virtual agents, they include many multi-modal factors such as appearance, while the scales are rather rough in terms of language operational ability.
Therefore, we discard, integrate and divide the original measures to fit our research objectives.
Our evaluation metrics are shown in \tabref{tab:criteria}.

\subsection{Collection of dialogue data}
\subsubsection{Participants}

In this study, we use crowdsourcing to conduct subjective evaluations.
We recruited 32 Japanese-speaking crowdworkers from a Japanese crowdsourcing service called Lancers\footnote{https://www.lancers.jp/}.
Only workers who performed high-quality work were selected.
The unit price was set at 300 yen (about three dollars) per dialogue task.
32 workers were assigned to all 25 systems and collected one dialogue for each system.

\subsubsection{Dialogue task settings}

In this study, based on a live dialogue system competition in Japan\cite{higashinaka2020dslc3}, the dialogue length was set to 15 turns each by the system and the user.
The conversation starts with a fixed phrase from the system: "Hello. Nice to meet you." 
After 15 utterances each, the following fixed phrases notify the user of the dialogue’s end: "Oh, I'm sorry. Our time is about up. Thank you for today." 
The "goodbye" answers the responses of the 15 user utterances, and the conversation is finished.
After each conversation, a link is displayed by Google Form that evaluates the conversation.
Interaction evaluations are done by a 11-stage Likert scale that ranges from 0 (completely disagree) to 10 (completely agree) for each item.

A telegram platform was used for the interactions. The dialog environment (PCs, smartphones, etc.) of the workers did not include any particular constraints.

\begin{table*}[h]
\begin{minipage}{0.5\linewidth}
    \centering
    \begin{tabular}{c|rrr|r}
Measure & ED & PC & Fav & Average\\\hline
Naturalness & 5.81$\uparrow$ & 5.00 & 6.41 & 5.74\\
Ease & 5.97 & 6.12 & 7.00 & 6.36\\
Enjoyment & 5.16 & 5.50 & 6.97 &5.88\\
Empathy & 4.25 & 4.94 & 6.03 & 5.07\\
Attentiveness & {\bf 4.31}$\downarrow$ & 5.34 & {\bf 8.12}$\uparrow$ & 5.93\\
Trust & 4.22 & 4.09 & 5.62 & 4.65\\
Personality & 5.53 & 5.19 & 6.09 & 5.60\\
Agency & 5.78 & 5.00 & 6.22 & 5.67\\
Topic & 5.03 & 5.38 & 7.03 & 5.81\\
Emotion & 5.53$\uparrow$ & 4.66 & 5.69$\downarrow$ & 5.29\\
Consistency & 4.41$\uparrow$ & 3.25 & 4.81 & 4.16\\
Engagement & 4.94 & 4.78 & 5.59 & 5.10\\
Respeak & 4.88 & 4.59 & 5.94 & 5.14\\\hline
Average & 5.06 & 4.91 & 6.27 & 
    \end{tabular}
    \subcaption{Human evaluations of models without additional information ({\it flat} condition)}
    \label{tab:eval_flat}
\end{minipage}
\begin{minipage}{0.5\linewidth}
    \centering
    \begin{tabular}{c|rrr|r}
Measure & ED & PC & Fav & Average\\\hline
Naturalness & 3.41 & 3.41$\downarrow$ & 7.09 &4.64\\
Ease & 4.12 & 3.81 & 7.12 &5.02\\
Enjoyment & 3.50 & 3.47 & 6.22$\downarrow$ &4.40\\
Empathy & 2.84 & 2.88 & 5.84 & 3.85\\
Attentiveness & {\bf 2.75}$\downarrow$ & 3.78 & 7.00 &4.51\\
Trust & 2.44 & 2.22 & 6.00 & 3.55\\
Personality & 3.66 & 3.53 & 6.06 &4.42\\
Agency & 4.12 & 3.47 & 6.50 &4.70\\
Topic & {\bf 5.00}$\uparrow$ & 5.03$\uparrow$ & 6.50 &5.51\\
Emotion & 2.97 & 3.44 & 5.91 &4.11\\
Consistency & 1.50$\downarrow$ & 2.81 & 5.91 &3.41\\
Engagement & 2.53 & 3.12 & 5.81 &3.82\\
Respeak & 2.78 & 3.00 & 5.88 &3.89\\\hline
Average & 3.20 & 3.38 & 6.30 
    \end{tabular}
    \subcaption{Human evaluations of models with additional information ({\it tagged} condition)}
    \label{tab:eval_tagged}
    \end{minipage}
    \caption{
    Human evaluations on multi-axial evaluation measures: Up arrows denote corresponding dataset significantly improved the evaluation metric, and down arrows denote decrease of metric (bold: $p<.05$, non-bold: $p<.1$).}

\end{table*}

\section{Results and analysis}

\subsection{Automatic evaluation}
\tabref{tab:pre-train} shows the perplexity of the pre-trained models on the validation set of pre-training dataset. The perplexity decrease with larger models. 

\tabref{tab:ppl} shows the perplexity of each model on the test set of each fine-tuning dataset.
For all fine-tuning datasets except Pre-trained, the larger the model size shows the lower the perplexity, and the use of additional information improves the perplexity.

\subsection{Human evaluation}

\subsubsection{User impressions of fine-tuning datasets}
We analyzed how the various datasets used for fine-tuning affected the user's impressions using a multi-axial evaluation scale.
\tabref{tab:eval_flat} shows the evaluation results using only the dataset sentences ({\it flat} condition).
\tabref{tab:eval_flat} shows that the ED dataset improved naturalness, emotion, and consistency but lowered attentiveness.
Since ED has empathetic utterances for many kinds of situations and emotions, the ED-trained model enables users to feel that the system shows natural and consistent emotions. However, since the dialogues in the ED dataset only included four utterances, the system barely developed dialogue topics and simply repeats empathetic utterances, which probably decreased attentiveness.
In contrast, \tabref{tab:eval_flat} illustrates that the PC has no significant effect on evoking specific impression including personality. 
\tabref{tab:eval_flat} also shows that Fav improved attentiveness scores but decreased emotion.
This is because the Fav dataset consists of long dialogues that contain many questions that effectively improve attentiveness.
On the other hand, such questions seem less effective to express the speaker's own emotions.
From the viewpoint of overall impression, the Fav dataset significantly outperforms the other two datasets. 


\tabref{tab:eval_tagged} shows the evaluation results that include additional information ({\it tagged} condition).
The main difference with the flat condition is the huge decrease of the average overall scores of ED and PC.
The ED and PC datasets have concrete dialogue topics defined with profile sentences or situation information.
Such information contributes to the improvement of topic scores, but in the actual dialogues, these systems with additional information frequently generate persistent utterances that have almost the same content of dialogue history.

\begin{figure}[t]
\centering
\includegraphics[width=.95\linewidth]{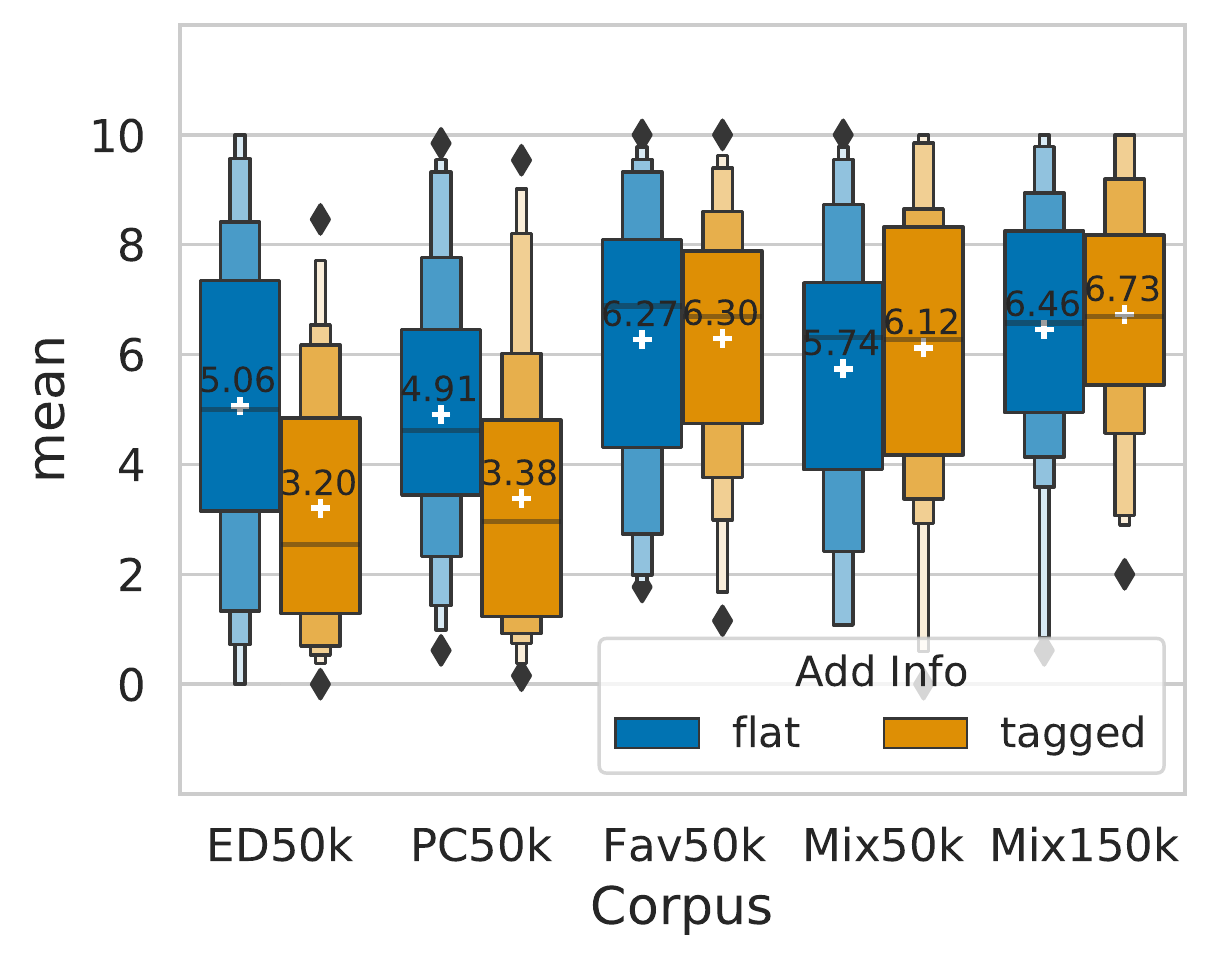}
\caption{Mixing datasets\label{fig:addinfo}}
\end{figure}

\begin{figure*}[t]
\includegraphics[width=1.05\linewidth]{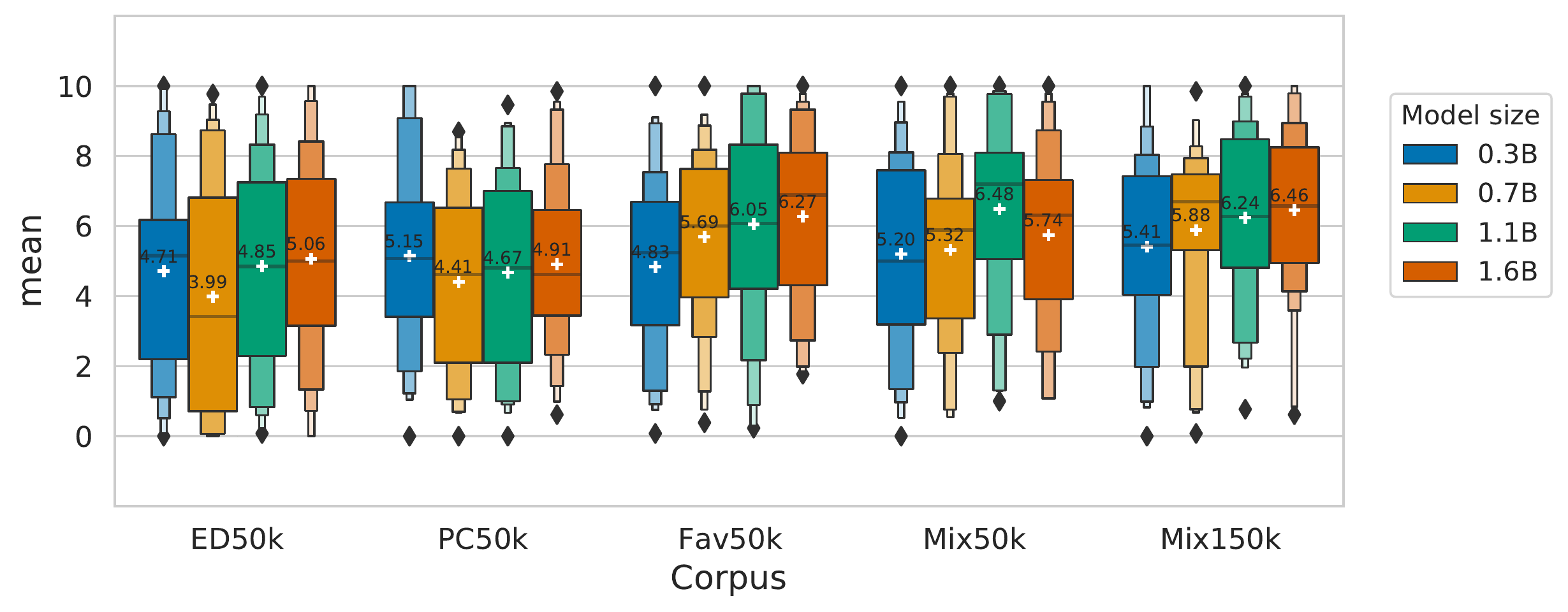}
\caption{Variation of model size and datasets\label{fig:datatype}}
\end{figure*}

\subsection{Mixing datasets}
We also tested whether mixing datasets improved the performance more than using a single dataset.
Figure \ref{fig:addinfo} shows the relationship between the corpus and the average evaluation value for each additional bit of information.
In all cases, Mix50K $<$ Fav50K $<$ Mix150K.
For the same amount of datasets, the performance is slightly degraded by mixing datasets with low evaluations, such as ED and PC.
On the other hand, when the total amount of datasets increased, the performance improved even when low evaluation datasets are mixed.

With respect to the presence or absence of additional information, the evaluation of single datasets (especially ED and PC) tended to decrease with additional information, and the performance of mixed datasets improved with additional information.
In the case of a mixture of different datasets, learning to distinguish the type and specifying it with the highest evaluation during inferences may have contributed to the performance improvement.

\subsubsection{Model size and datasets}

The performance of the model size for each corpus is shown in Figure \ref{fig:datatype}.
For Fav50k and Mix150k, the model size is significantly correlated with the evaluation value.
On the other hand, for ED50k, PC50k, and Mix50k, the correlation between the model size and evaluation value was insignificant, indicating that the evaluation did not increase with the model size.
In general, performance improved as model size increased, and in fact, the perplexity improved with increasing model size for all the models in Mix and Fav, although the results were different from the objective evaluation. In fact, the perplexity improved with increasing model size for all models in both Mix and Fav; the results were different from the objective evaluations. This suggests that in long dialogues, factors that strongly affect the impression cannot be measured by simple perplexity.





\section{Conclusion}
We developed the largest Transformer-based Japanese dialogue models and Japanese version of PersonaChat and EmpatheticDialogues, which are widely used standard benchmarking dataset for evaluating chit-chat models in English.
We also evaluated and analyzed the effects of the changes in the fine-tuning datasets, model size, and the use of additional information on users' impressions of dialogues from multiple perspectives.
Our results identified the following: The model performance and user impressions greatly depend on the sentences contained in fine-tuning dataset, and this effect exists when additional information (e.g., profile sentences) is not available.
The use of additional information is intended to improve specific impressions, but it is not always beneficial.
A relationship between model size and overall performance varies greatly depending on the type of fine-tuned corpus. We found that the model size did not improve the PersonaChat and EmpatheticDialogues performance.

Future work will clarify the relationship how dialogue content or degrees of breakdown affect to dialogue impressions.

\section*{Acknowledgement}
This work was supported by JSPS KAKENHI Grant Number 19H05693.

\bibliographystyle{acl_natbib}
\bibliography{anthology,acl2021,references}


\end{document}